\documentclass[conference]{IEEEtran}
\usepackage{graphicx}
\usepackage{subfig, url}
\usepackage{amsmath, amssymb, bbm, nicefrac}
\usepackage{hyperref}
\usepackage{float}
\usepackage{dirtytalk}

\title{
    Real-time Convolutional Neural Networks for Emotion and Gender Classification
}

\author{\IEEEauthorblockN{Octavio Arriaga}
        \IEEEauthorblockA{Hochschule Bonn-Rhein-Sieg \\
        Sankt Augustin Germany  \\
        Email: octavio.arriaga@smail.inf.h-brs.de}
        \and
        \IEEEauthorblockN{Paul G. Pl\"oger}
        \IEEEauthorblockA{Hochschule Bonn-Rhein-Sieg \\ 
        Sankt Augustin Germany \\
        Email: paul.ploeger@h-brs.de}
        \and
        \IEEEauthorblockN{Matias Valdenegro}
        \IEEEauthorblockA{ Heriot-Watt University \\
        Edinburgh, UK \\
        Email: m.valdenegro@hw.ac.uk}}

\begin{document}

    \maketitle
    \begin{abstract}


        In this paper we propose an implement a general convolutional neural network (CNN) building framework for designing real-time CNNs.
        We validate our models by creating a real-time vision system which accomplishes the tasks of face detection, gender classification and emotion classification simultaneously in one blended step using our proposed CNN architecture.
        After presenting the details of the  training procedure setup we proceed to evaluate on standard benchmark sets.
        We report accuracies of 96\% in the IMDB gender dataset and 66\% in the FER-2013 emotion dataset.
        Along with this we also introduced the very recent real-time enabled guided back-propagation visualization technique.
        Guided back-propagation uncovers the dynamics of the weight changes and evaluates the learned features.
        We argue that the careful implementation of modern CNN architectures, the use of the current regularization methods and the visualization of previously hidden features are necessary in order to reduce the gap between slow performances and real-time architectures.
        Our system has been validated by its deployment on a Care-O-bot 3 robot used during RoboCup@Home competitions.
        All our code, demos and pre-trained architectures have been released under an open-source license in our \href{https://github.com/oarriaga/face_classification/tree/master}{public repository}.
    \end{abstract}

    \section{Introduction}

\begin{figure}
        \centering
        \includegraphics[scale=.31]{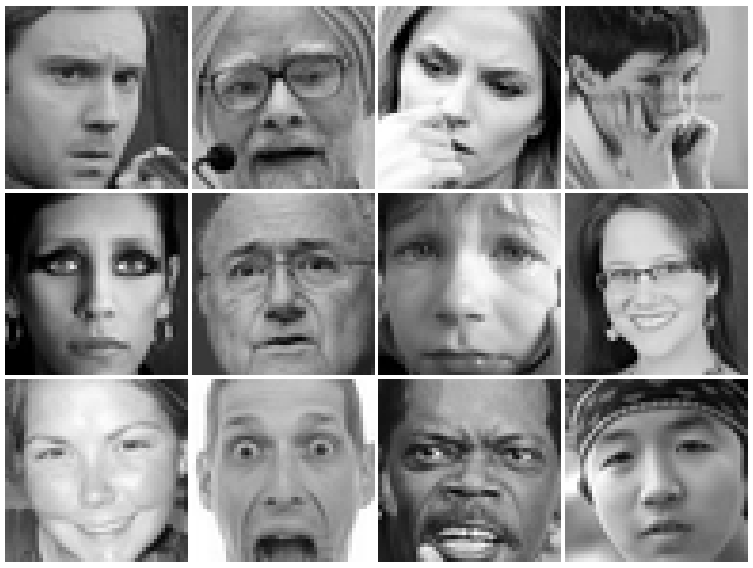}
        \caption{Samples of the FER-2013 emotion dataset \cite{Goodfellow-2013}.}
        \label{fig:emotion_dataset}
    \end{figure}

    The success of service robotics decisively depends on a smooth robot to user interaction. Thus, a robot should be able to extract information just from the face of its user, e.g. identify the emotional state or deduce gender.
    Interpreting correctly any of these elements using machine learning (ML) techniques has proven to be complicated due the high variability of the samples within each task \cite{Goodfellow-2013}.
    This leads to models with millions of parameters trained under thousands of samples \cite{Dario-2015}.
    Furthermore, the human accuracy for classifying an image of a face in one of 7 different emotions is 65\% $\pm$ 5\% \cite{Goodfellow-2013}.
    One can observe the difficulty of this task by trying to manually classify the FER-2013 dataset images in Figure \ref{fig:emotion_dataset} within the following classes \{\say{angry}, \say{disgust}, \say{fear}, \say{happy}, \say{sad}, \say{surprise}, \say{neutral}\}.

    \begin{figure}
        \centering
        \includegraphics[scale=.46]{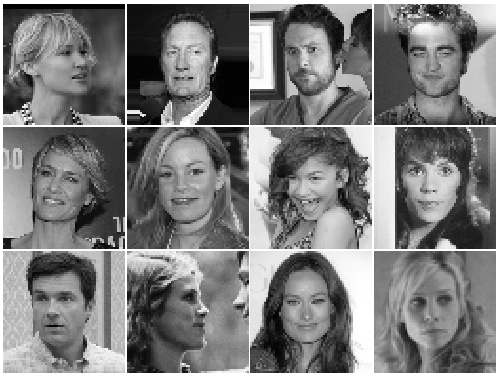}
        \caption{Samples of the IMDB dataset \cite{Rothe-IJCV-2016}.}
        \label{fig:gender_dataset}
    \end{figure}

 
    In spite of these difficulties, robot platforms oriented to attend and solve household tasks require facial expressions systems that are robust and computationally efficient.
    Moreover, the state-of-the-art methods in image-related tasks such as image classification \cite{Chollet-2016} and object detection are all based on Convolutional Neural Networks (CNNs).
    These tasks require CNN architectures with millions of parameters; therefore, their deployment in robot platforms and real-time systems becomes unfeasible.
    In this paper we propose an implement a general CNN building framework for designing real-time CNNs.
    The implementations have been validated in a real-time facial expression system that provides face-detection, gender classification and that achieves human-level performance when classifying emotions.
    This system has been deployed in a care-O-bot 3 robot, and has been extended for general robot platforms and the RoboCup@Home competition challenges.

    Furthermore, CNNs are used as black-boxes and often their learned features remain hidden, making it complicated to establish a balance between their classification accuracy and unnecessary parameters.
    Therefore, we implemented a real-time visualization of the guided-gradient back-propagation proposed by Springenberg \cite{Springenberg-2014} in order to validate the features learned by the CNN.

    \section{Related Work}

    Commonly used CNNs for feature extraction include a set of fully connected layers at the end. 
    Fully connected layers tend to contain most of the parameters in a CNN.
    Specifically, VGG16 \cite{Simonyan2014} contains approximately 90\% of all its parameters in their last fully connected layers.
    Recent architectures such as Inception V3 \cite{Szegedy2016rethinking}, reduced the amount of parameters in their last layers by including a Global Average Pooling operation.
    Global Average Pooling reduces each feature map into a scalar value by taking the average over all elements in the feature map.
    The average operation forces the network to extract global features from the input image.
    Modern CNN architectures such as Xception \cite{Chollet-2016} leverage from the combination of two of the most successful experimental assumptions in CNNs: the use of residual modules \cite{He-2016} and depth-wise separable convolutions \cite{Howard-2017}.
    Depth-wise separable convolutions reduce further the amount of parameters by separating the processes of feature extraction and combination within a convolutional layer.

    Furthermore, the state-of-the-art model for the FER2-2013 dataset is based on CNN trained with square hinged loss \cite{tang2013deep}. 
    This model achieved an accuracy of 71\% \cite{Goodfellow-2013} using approximately 5 million parameters.
    In this architecture 98\% of all parameters are located in the last fully connected layers.
    
    The second-best methods presented in \cite{Goodfellow-2013} achieved an accuracy of 66\% using an ensemble of CNNs.

    \section{Model}
    We propose two models which we evaluated in accordance to their test accuracy and number of parameters.
    Both models were designed with the idea of creating the best accuracy over number of parameters ratio.
    Reducing the number of parameters help us overcoming two important problems.
    First, the use of small CNNs alleviate us from slow performances in hardware-constrained systems such robot platforms.
    And second, the reduction of parameters provides a better generalization under an Occam's razor framework.
    Our first model relies on the idea of eliminating completely the fully connected layers.
    The second architecture combines the deletion of the fully connected layer and the inclusion of the combined depth-wise separable convolutions and residual modules.
    Both architectures were trained with the ADAM optimizer \cite{kingma2014adam}.

    Following the previous architecture schemas, our initial architecture used Global Average Pooling to completely remove any fully connected layers.
    This was achieved by having in the last convolutional layer the same number of feature maps as number of classes, and applying a softmax activation function to each reduced feature map.
    Our initial proposed architecture is a standard fully-convolutional neural network composed of 9 convolution layers, ReLUs \cite{Glorot-2011}, Batch Normalization \cite{Ioffe-2015} and Global Average Pooling.
    This model contains approximately 600,000 parameters.
    It was trained on the IMDB gender dataset, which contains 460,723 RGB images where each image belongs to the class \say{woman} or \say{man}, and it achieved an accuracy of 96\% in this dataset.
    We also validated this model in the FER-2013 dataset.
    This dataset contains 35,887 grayscale images where each image belongs to one of the following classes \{\say{angry}, \say{disgust}, \say{fear}, \say{happy}, \say{sad}, \say{surprise}, \say{neutral}\}.
    Our initial model achieved an accuracy of 66\% in this dataset.
    We will refer to this model as \say{sequential fully-CNN}.

    \begin{figure}
        \centering
        \includegraphics[scale=.5]{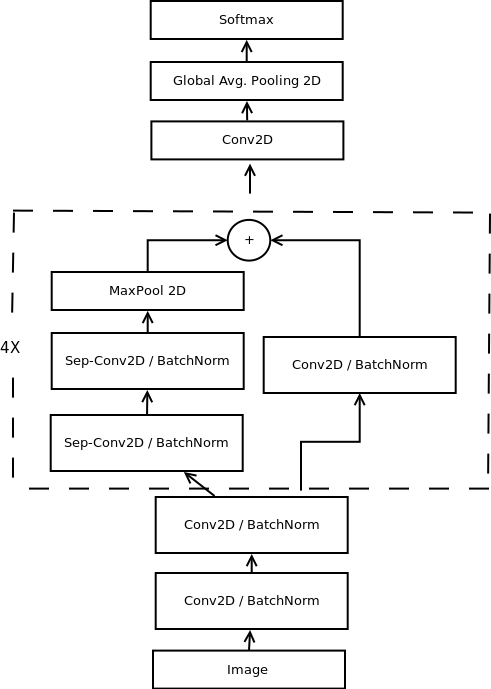}
        \caption{Our proposed model for real-time classification.}
        \label{fig:mini_xception}
    \end{figure}

    Our second model is inspired by the Xception \cite{Chollet-2016} architecture.
    This architecture combines the use of residual modules \cite{He-2016} and depth-wise separable convolutions \cite{Howard-2017}.
    Residual modules modify the desired mapping between two subsequent layers, so that the learned features become the difference of the original feature map and the desired features.
    Consequently, the desired features $H(x)$ are modified in order to solve an easier learning problem $F(X)$ such that:
    \begin{equation}
        H(x) = F(x) + x
    \end{equation}

    Since our initial proposed architecture deleted the last fully connected layer, we reduced further the amount of parameters by eliminating them now from the convolutional layers.
    This was done trough the use of depth-wise separable convolutions.
    Depth-wise separable convolutions are composed of two different layers: depth-wise convolutions and point-wise convolutions.
    The main purpose of these layers is to separate the spatial cross-correlations from the channel cross-correlations \cite{Chollet-2016}.
    They do this by first applying a $D \times D$ filter on every $M$ input channels and then applying $N$ $1 \times 1 \times M$ convolution filters to combine the $M$ input channels into $N$ output channels.
    Applying $1 \times 1 \times M$ convolutions combines each value in the feature map without considering their spatial relation within the channel.

    \begin{figure}
            \centering
                \subfloat[]{\label{fig:standard_convolution}
                \includegraphics[scale=.4]{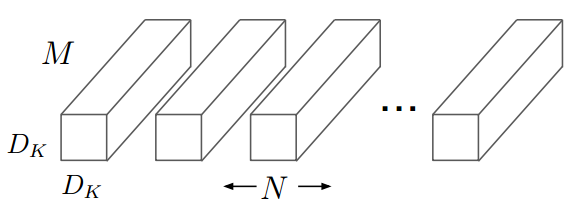}}
                \hspace{0.5cm}    
                \subfloat[]{\label{fig:depth_wise_separable_convolution}
                \includegraphics[scale=.3]{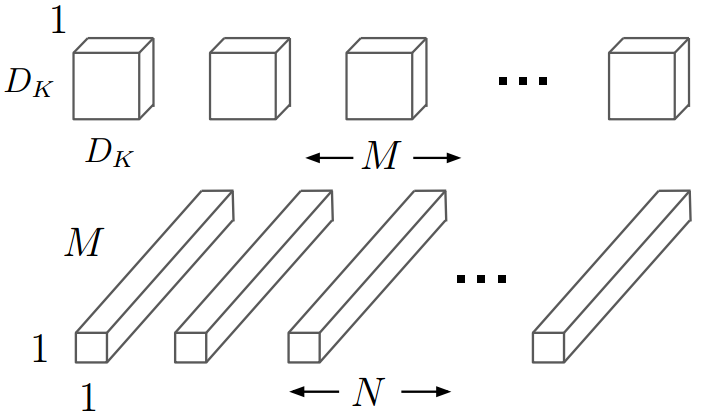}}
                \caption{\cite{Howard-2017} Difference between (a) standard convolutions and (b) depth-wise separable convolutions.}
            \label{fig:convolutions}
        \end{figure}

    Depth-wise separable convolutions reduces the computation with respect to the standard convolutions by a factor of $\frac{1}{N} + \frac{1}{D^{2}}$ \cite{Howard-2017}.
    A visualization of the difference between a normal Convolution layer and a depth-wise separable convolution can be observed in Figure \ref{fig:convolutions}.

    Our final architecture is a fully-convolutional neural network that contains 4 residual depth-wise separable convolutions where each convolution is followed by a batch normalization operation and a ReLU activation function.
    The last layer applies a global average pooling and a soft-max activation function to produce a prediction.
    This architecture has approximately $60,000$ parameters; which corresponds to a reduction of $10 \times$ when compared to our initial naive implementation, and $80 \times$ when compared to the original CNN.
    Figure \ref{fig:mini_xception} displays our complete final architecture which we refer to as mini-Xception.
    This architectures obtains an accuracy of $95\%$ in gender classification task.
    Which corresponds to a reduction of one percent with respect to our initial implementation.
    Furthermore, we tested this architecture in the FER-2013 dataset and we obtained the same accuracy of $66\%$ for the emotion classification task.
    Our final architecture weights can be stored in an 855 kilobytes file.
    By reducing our architectures computational cost we are now able to join both models and use them consecutively in the same image without any serious time reduction.
    Our complete pipeline including the openCV face detection module, the gender classification and the emotion classification takes $0.22 \pm 0.0003$ ms on a i5-4210M CPU.
    This corresponds to a speedup of $1.5 \times$ when compared to the original architecture of Tang.

    We also added to our implementation a real-time guided back-propagation visualization to observe which pixels in the image activate an element of a higher-level feature map.
    Given a CNN with only ReLUs as activation functions for the intermediate layers, guided-back propagation takes the derivative of every element $(x, y)$ of the input image $I$ with respect to an element $(i, j)$ of the feature map $f^{L}$ in layer $L$.
    The reconstructed image $R$ filters all the negative gradients; consequently, the remaining gradients are chosen such that they only increase the value of the chosen element of the feature map.
    Following \cite{Springenberg-2014}, a fully ReLU CNN reconstructed image in layer $l$ is given by:
    \begin{equation}
            R_{i,j}^l = (R_{i,j}^{l+1} > 0) * R_{i,j}^{l + 1}
    \end{equation}

    \section{Results}
\begin{figure}
        \centering
        \includegraphics[scale=.24]{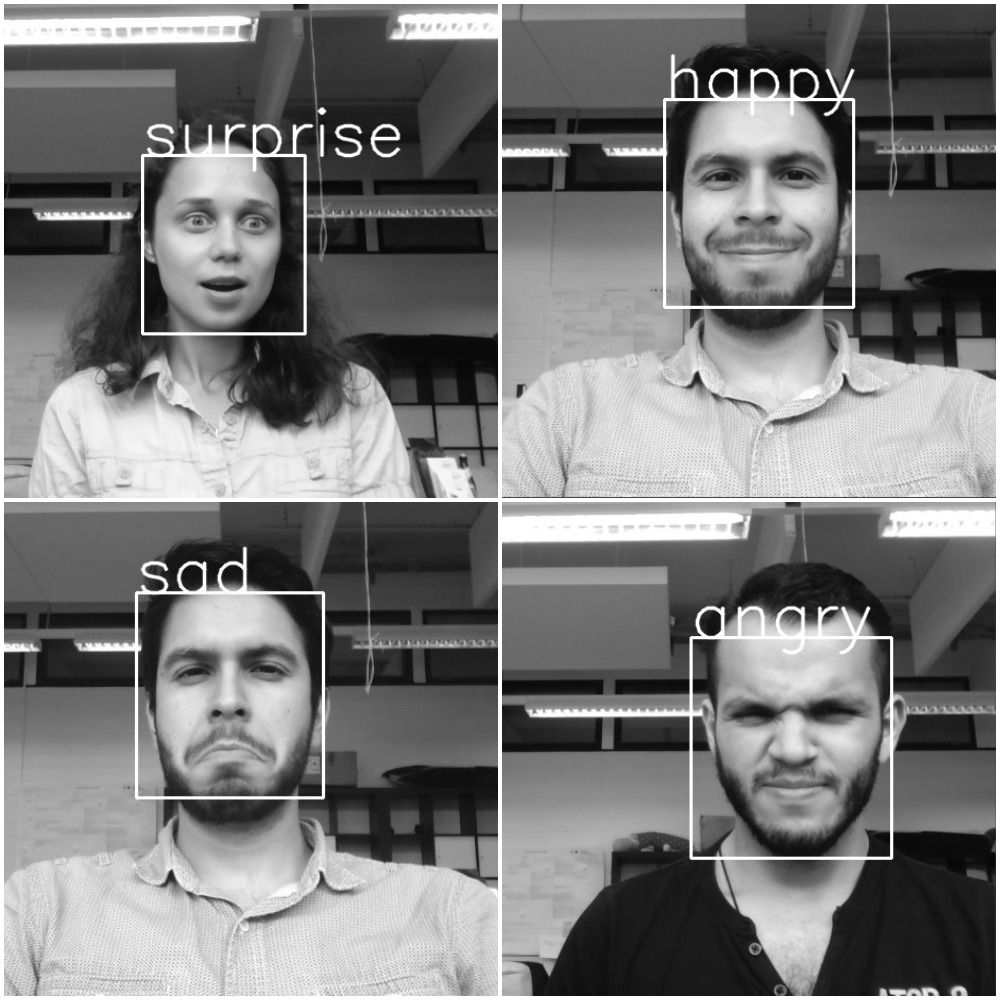}
        \caption{Results of the provided real-time emotion classification provided in our public repository}
        \label{fig:emotion_demo}
    \end{figure}

    Results of the real-time emotion classification task in unseen faces can be observed in Figure \ref{fig:emotion_demo}. 
    Our complete real-time pipeline including: face detection, emotion and gender classification have been fully integrated in our Care-O-bot 3 robot.

    \begin{figure}
        \centering
        \includegraphics[scale=.25]{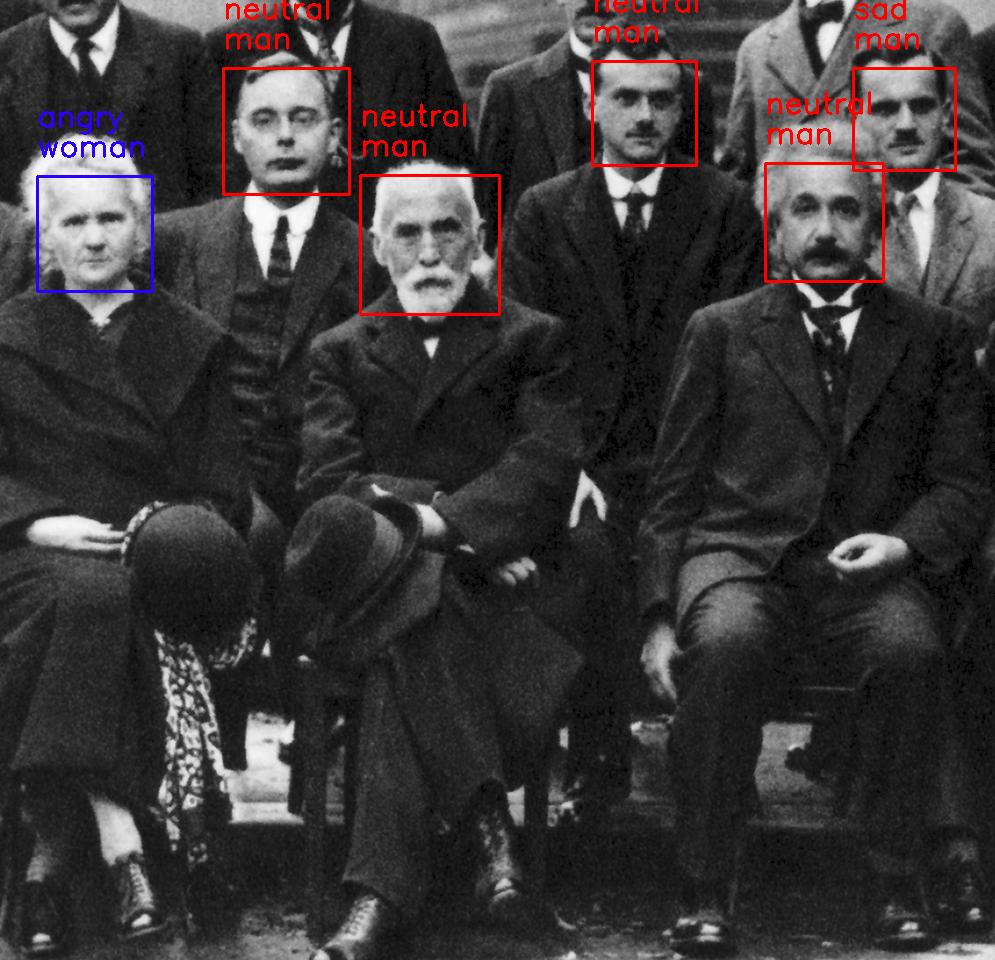}
        \caption{Results of the provided combined gender and emotion inferences demo. The color blue represents the assigned class \textit{woman} and red the class \textit{man}}
        \label{fig:gender_demo}
    \end{figure}

\begin{figure}
        \centering
        \includegraphics[scale=.85]{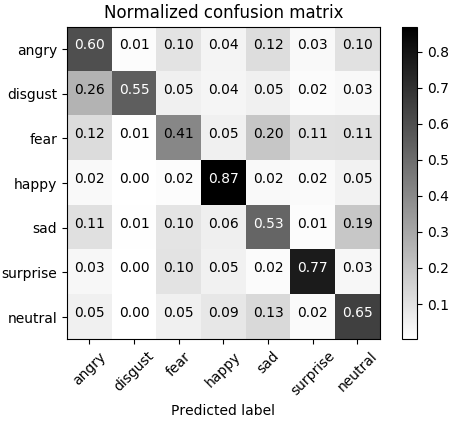}
        \caption{Normalized confusion matrix of our mini-Xception network.}
        \label{fig:confusion_matrix}
    \end{figure}

    An example of our complete pipeline can be seen in Figure \ref{fig:gender_demo} in which we provide emotion and gender classification.
    In Figure \ref{fig:confusion_matrix} we provide the confusion matrix results of our emotion classification mini-Xception model.
    We can observe several common misclassifications such as predicting \say{sad} instead of \say{fear} and predicting \say{angry} instead \say{disgust}.

\begin{figure}
        \centering
        \subfloat[]{\label{fig:gbp_1}
        \includegraphics[scale=.4]{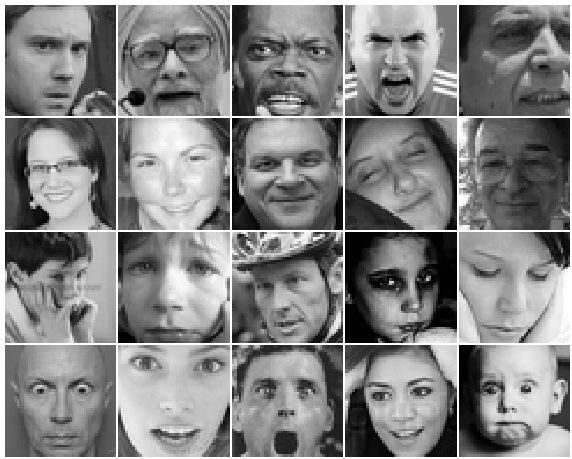}}
        \hspace{0.5cm}    
        \subfloat[]{\label{fig:gbp_2}
        \includegraphics[scale=.4]{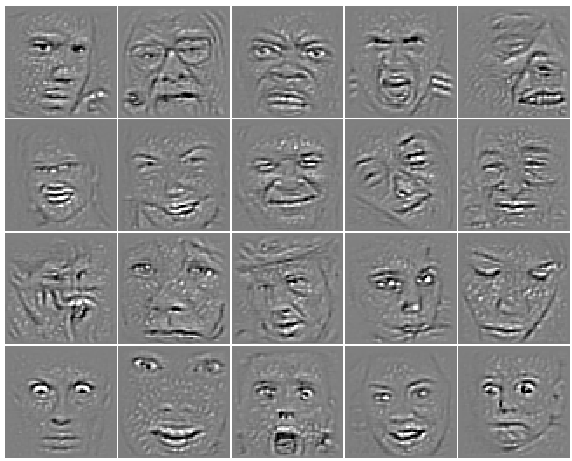}}
        \hspace{0.5cm}    
        \subfloat[]{\label{fig:gbp_3}
        \includegraphics[scale=.48]{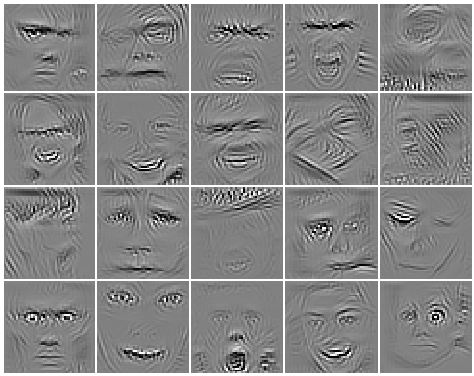}}
        \caption{All sub-figures contain the same images in the same order. Every row starting from the top corresponds respectively to the emotions: angry, happy, sad and surprise (a) Samples from the FER-2013 dataset (b) Guided back-propagation visualization of our mini-Xception model (c) Guided back-propagation visualization of our sequential fully-CNN.}
        \label{fig:gbp}
    \end{figure}

    A comparison of the learned features between several emotions and both of our proposed models can be observed in Figure \ref{fig:gbp}.
    The white areas in figure \ref{fig:gbp_2} correspond to the pixel values that activate a selected neuron in our last convolution layer.
    The selected neuron was always selected in accordance to the highest activation.
    We can observe that the CNN learned to get activated by considering features such as the frown, the teeth, the eyebrows and the widening of one's eyes, and that each feature remains constant within the same class.
    These results reassure that the CNN learned to interpret understandable human-like features, that provide generalizable elements.
    These interpretable results have helped us understand several common misclassification such as persons with glasses being classified as \say{angry}.
    This happens since the label \say{angry} is highly activated when it believes a person is frowning and frowning features get confused with darker glass frames.
    Moreover, we can also observe that the features learned in our mini-Xception model are more interpretable than the ones learned from our sequential fully-CNN.
    Consequently the use of more parameters in our naive implementations leads to less robust features.

    \section{Future work}

    Machine learning models are biased in accordance to their training data.
    In our specific application we have empirically found that our trained CNNs for gender classification are biased towards western facial features and facial accessories.
    We hypothesize that this misclassfications occurs since our training dataset consist of mostly western: actors, writers and cinematographers as observed in Figure \ref{fig:gender_dataset}.

    Furthermore, as discussed previously, the use of glasses might affect the emotion classification by interfering with the features learned. However, the use of glasses can also interfere with the gender classification.
    This might be a result from the training data having most of the images of persons wearing glasses assigned with the label \say{man}.
    We believe that uncovering such behaviours is of extreme importance when creating robust classifiers, and that the use of the visualization techniques such as guided back-propagation will become invaluable when uncovering model biases.

    \newpage
    \section{Conclusions}

    We have proposed and tested a general building designs for creating real-time CNNs.
    Our proposed architectures have been systematically built in order to reduce the amount of parameters.
    We began by eliminating completely the fully connected layers and by reducing the amount of parameters in the remaining convolutional layers via depth-wise separable convolutions.
    We have shown that our proposed models can be stacked for multi-class classifications while maintaining real-time inferences.
    Specifically, we have developed a vision system that performs face detection, gender classification and emotion classification in a single integrated module.
    We have achieved human-level performance in our classifications tasks using a single CNN that leverages modern architecture constructs.
    Our architecture reduces the amount of parameters $80 \times$ while obtaining favorable results.
    Our complete pipeline has been successfully integrated in a Care-O-bot 3 robot.
    Finally we presented a visualization of the learned features in the CNN using the guided back-propagation visualization.
    This visualization technique is able to show us the high-level features learned by our models and discuss their interpretability.

    \section*{Acknowledgments}
    We gratefully acknowledge the continued support by the b-it Bonn-Aachen International Center for Information Technology and the Hochschule Bonn-Rhein-Sieg.

    \bibliographystyle{plain}
    \bibliography{references}

\end{document}